\documentclass{article}

\usepackage[main,final]{neurips_2026}
\usepackage{algorithm}
\usepackage{algorithmic}
\usepackage{graphicx}
\usepackage{enumitem}
\usepackage{tcolorbox}
\usepackage{wrapfig}
\usepackage{graphicx}
\usepackage{array}
\usepackage{booktabs}
\usepackage[table]{xcolor}
\usepackage{siunitx}
\usepackage{subcaption}
\usepackage{xcolor}
\usepackage{float}
\definecolor{stdgray}{gray}{0.35}
\newcommand{\stderr}[1]{\textcolor{stdgray}{\scriptsize\,{\(\pm #1\)}}}
\definecolor{mainblue}{RGB}{235,245,255}
\newcommand{\sigp}[1]{\cellcolor{green!25}{#1}}
\usepackage{xcolor}

\usepackage{multirow}

\usepackage[utf8]{inputenc} 
\usepackage[T1]{fontenc}    
\usepackage{hyperref}       
\usepackage{url}            
\usepackage{booktabs}       
\usepackage{amsfonts}       
\usepackage{nicefrac}       
\usepackage{microtype}      
\usepackage{xcolor}         

\usepackage{nicematrix}
\usepackage{xspace}

\usepackage{amsmath,amsfonts,bm}

\def\eqref#1{equation~\ref{#1}}

\def\1{\bm{1}}

\DeclareMathAlphabet{\mathsfit}{\encodingdefault}{\sfdefault}{m}{sl}
\SetMathAlphabet{\mathsfit}{bold}{\encodingdefault}{\sfdefault}{bx}{n}

\def\gP{{\mathcal{P}}}

\newcommand{\method}{DyStruct\xspace}

\title{\method: Dynamically Structured Diffusion Language Model Decoding via Bayesian Inference}

\author{
  Bian Sun\thanks{Equal contribution.}  \\
    Institute for Artificial Intelligence\\
  University of Central Florida\\
  \texttt{bian.sun@ucf.edu} \\
  \And
  Kevin Zhai\footnotemark[1] \\
      Institute for Artificial Intelligence\\
  University of Central Florida\\
  \texttt{Kevin.Zhai@ucf.edu} \\
  \And
  Mubarak Shah \\
      Institute for Artificial Intelligence\\
  University of Central Florida\\
  \texttt{shah@crcv.ucf.edu} \\
  \And
  Zhenyi Wang\thanks{Corresponding author} \\
      Institute for Artificial Intelligence\\
  University of Central Florida\\
  \texttt{zhenyi.wang@ucf.edu} \\
}

\begin{document}

\maketitle

\begin{abstract}
Diffusion language models (DLMs) have recently emerged as a promising alternative to autoregressive models, primarily due to their ability to enable parallel decoding. Despite this advantage, most existing DLMs rely on a fixed generation length specified prior to decoding, which restricts their flexibility in real-world applications. While a few recent works attempt to support flexible-length generation, they typically suffer from notable limitations: some require costly retraining to accommodate variable-length outputs, while others depend solely on local confidence signals during decoding. Such local criteria fail to capture the evolving structure of the sequence, often resulting in suboptimal generation quality.
In this paper, we propose a training-free, Bayesian structured decoding framework that formulates flexible-length generation as a dynamic structural inference problem. Our approach learns the posterior inference over the dynamic block length, block formations and growth, and block decoding order within a unified Bayesian inference framework to jointly reason about how much to grow, where to grow,  and how to organize content structurally. At each window expansion step, the method integrates local uncertainty with structural signals to (i) dynamically expand the sequence via adaptive length growth, (ii) infer block boundaries through Chinese Restaurant Process (CRP)-style partitioning, and (iii) allocate different number of decoding steps for different blocks and determine block decoding order via context-aware scheduling. This yields a unified mechanism that supports dynamic structured generation, including both flexible block expansion and block organization, while maintaining coherence.
Extensive experiments across multiple benchmarks demonstrate that our approach significantly improves generation quality and flexibility over existing fixed-length and flexible-length baselines. These results highlight the advantage of Bayesian structured decoding for diffusion language model, providing a principled and efficient solution for structured text generation. 
\end{abstract}
\section{Introduction}

Most large language models (LLMs) \citep{NEURIPS2020_1457c0d6} rely on autoregressive decoding, where tokens are generated sequentially. This process limits decoding efficiency, especially for long sequences, because each new token depends on all previously generated tokens and cannot be produced in parallel. As a result, autoregressive LLMs often suffer from high inference latency and computational cost during deployment. Diffusion language models (DLMs) \citep{sahoo2024simple, nie2025large} offer an efficient alternative by enabling parallel decoding. Instead of predicting tokens one by one, diffusion-based approaches iteratively refine multiple token positions simultaneously. This parallel decoding paradigm makes diffusion language models a promising direction for building faster and scalable language generation systems.

However, DLMs typically rely on a fixed, pre-specified generation length. This assumption restricts practical flexibility, as the optimal output length depends on task complexity: complex queries require detailed responses, whereas simpler inputs call for concise outputs. Consequently, fixed-length decoding leads to either truncation or redundancy. More critically, it prevents the model from adapting generation to the evolving semantic context, highlighting the need for mechanisms that dynamically adjust sequence length during generation.

Several recent works attempt to relax this fixed-length assumption in DLMs, but existing approaches exhibit key limitations. FlexMDM \citep{kim2026anyorder} and DID \citep{ding2026beyond} rely on retraining to enable variable-length decoding, which incurs substantial computational cost. DAEDAL \citep{li2026beyond} avoids retraining but depends on heuristic, local confidence-based criteria. Crucially, these approaches overlook content organization after sequence expansion. When new tokens are generated, the lack of structural guidance results in fragmented structure.

To overcome these limitations, we formulate variable-length generation as structured decoding via Bayesian inference. We model the joint posterior distribution over the new window expansion size, the partition of the window into contiguous blocks, and the block decoding schedule. We introduce a structured prior over the latent block partition to govern content organization and guide decoding. This prior encourages coherent partition patterns while avoiding rigid assumptions about the number or boundaries of blocks. Specifically, we model block formation through a Chinese Restaurant Process (CRP) \citep{blei2010nested}. The advantages of the CRP are threefold: (1) it removes the need to preset the number of blocks, letting the model determine the quantity adaptively; (2) it does not require predefined partition boundaries, allowing the model to infer splits directly from the data; and (3) it provides a predictive prior over the partition structure, allowing decoding to assess at each step whether the current token should continue the current block or initiate a new one. This framework provides a training-free mechanism for sequence growth, allowing the model to jointly determine how much content to introduce, where to expand, and how newly generated tokens should be organized into contiguous blocks. By unifying local evidence with structural constraints, our approach enables flexible, coherent decoding without modifying the underlying model parameters. The overall algorithm is illustrated in Figure \ref{fig:overview}.

We evaluate this framework across diverse language generation tasks. The results demonstrate that performing joint structural inference at decoding time actively prevents sequence fragmentation, improving both generation quality and coherence while keeping the model completely frozen.

Our contributions are summarized as follows:
\begin{itemize}[leftmargin=0.5cm]
    \item We introduce a training-free Bayesian framework for dynamic structured decoding in diffusion language models, formulating flexible-length generation as joint inference over the new window size, block partitions, and decoding organization.
    \item We develop an efficient posterior inference algorithm to estimate dynamic window expansion, block partitioning via the CRP, and blocks decoding scheduling via context-aware prioritization.
    \item We evaluate the method across multiple datasets, demonstrating improved flexible-length generation quality and coherence without additional training.
\end{itemize}
\section{Related Work}

\textbf{Diffusion Language Models (DLMs).}
Recent advances establish DLMs by applying denoising diffusion probabilistic models \citep{ho2020denoising} through masked discrete formulations, improved training objectives and large-scale pretrained models. These developments demonstrate that diffusion is both a controllable generation framework and a viable foundation-modeling paradigm for language \citep{hoogeboom2021argmax,li2022diffusion, yu2022latent, savinov2022stepunrolled, reid2023diffuser, gulrajani2023likelihoodbased,he2023diffusionbert, gong2023diffuseq, lovelace2023latent, gat2024discrete, sahoo2024simple, lou2024discrete, liu2024unified, shi2024simplified, nie2025scaling, nie2025large, liu2025discrete, ye2025beyond, xu2025energybased, gong2025scaling, deschenaux2025beyond, rutte2025generalized, liu2025think, arriola2025block, zheng2025masked, sahoo2025the, zhang2025target, kim2025train, rout2025anchored, seo2025fast}. Despite this empirical success, DLMs share two practical limitations. First, most existing approaches operate under a fixed-length decoding setting, restricting real-world applicability where generation length must adapt to task complexity. Second, parallel generation in these models introduces distributional drift due to the conditional independence assumption across tokens \citep{guo2026selfspeculative}. Recent strategies like Hierarchy-dLLM \citep{qi2026hierarchy} attempt to mitigate this drift via hierarchical decoding. However, it relies on heuristic, position-based rules under a \textit{fixed-length} setting and lacks explicit modeling of decoding structure or content planning. In contrast, we formulate decoding as a Bayesian structural inference problem, jointly inferring new window size, block partitioning, and decoding order within a unified probabilistic framework. This framework enables dynamic length adaptation and coherent content organization, moving beyond local spatial heuristics toward structure-aware generation.

\textbf{Variable-Length Generation in DLMs.}
Recent works attempt to relax the fixed-length constraint, but existing approaches exhibit key limitations. DID \citep{ding2026beyond} and FlexMDM \citep{kim2026anyorder} enable dynamic token adjustments during generation, but lack an explicit model of decoding structure and require extensive retraining or alterations to the forward process. DAEDAL \citep{li2026beyond} avoids retraining but relies on heuristic, local confidence-based rules to expand the sequence. By contrast, our training-free approach replaces local heuristics with a joint inference framework, allowing the frozen model to determine where to expand, how much to expand, and how new content organizes into contiguous blocks.
\section{Problem Formulation}

We formulate the structured decoding problem. DLMs generate sequences through iterative refinement. Let $x$ denote the input prompt and let $\mathcal V$ denote the vocabulary. We define $t \in \{1, \dots, T_{ext}\}$ as the expansion step index. After step $t-1$, the current response is $y^{(t-1)}=(y^{(t-1)}_1,\dots,y^{(t-1)}_{n_{t-1}})$, where each position is either a vocabulary token or \texttt{[MASK]}. To predict the masked values, the model processes the concatenated sequence $[x; y^{(t-1)}]$. To enable flexible-length generation, the decoder appends a new masked window of length $L_t$ at expansion step $t$:
$
\tilde y^{(t)}
=
\bigl[
y^{(t-1)};
\underbrace{\texttt{[MASK]},\dots,\texttt{[MASK]}}_{L_t\ \text{new positions}}
\bigr].
$
We index positions inside this newly appended window locally by $j=1,\dots,L_t$, corresponding to global indices $n_{t-1}+j$ in $\tilde y^{(t)}$. Appending this window introduces a structural inference problem. At each expansion step $t$, the decoder needs to infer: (i) the allocated window length $L_t$, (ii) a partition $\mathcal P^{(t)}=\{B_1^{(t)}, \dots, B_{M_t}^{(t)}\}$ dividing the window into $M_t$ contiguous blocks, and (iii) a schedule $\tau^{(t)}$, which is a permutation of $\{1, \dots, M_t\}$ denoting the decoding order. To determine the partition $\mathcal P^{(t)}$, the decoder utilizes a CRP prior, governed by a concentration parameter $\alpha$, to evaluate whether adjacent positions should extend an existing block or initialize a new block. Once the structure $(\mathcal{P}^{(t)}, \tau^{(t)})$ is established, the decoder decodes each block through a series of unmasking iterations, the total count of which is dynamically determined based on block instability. We detail the notation in Appendix Table \ref{tab:notation}.

\section{Method}
\label{sec:method}

\begin{figure}[t]
\centering
\includegraphics[width=\linewidth ,height=9.5cm]{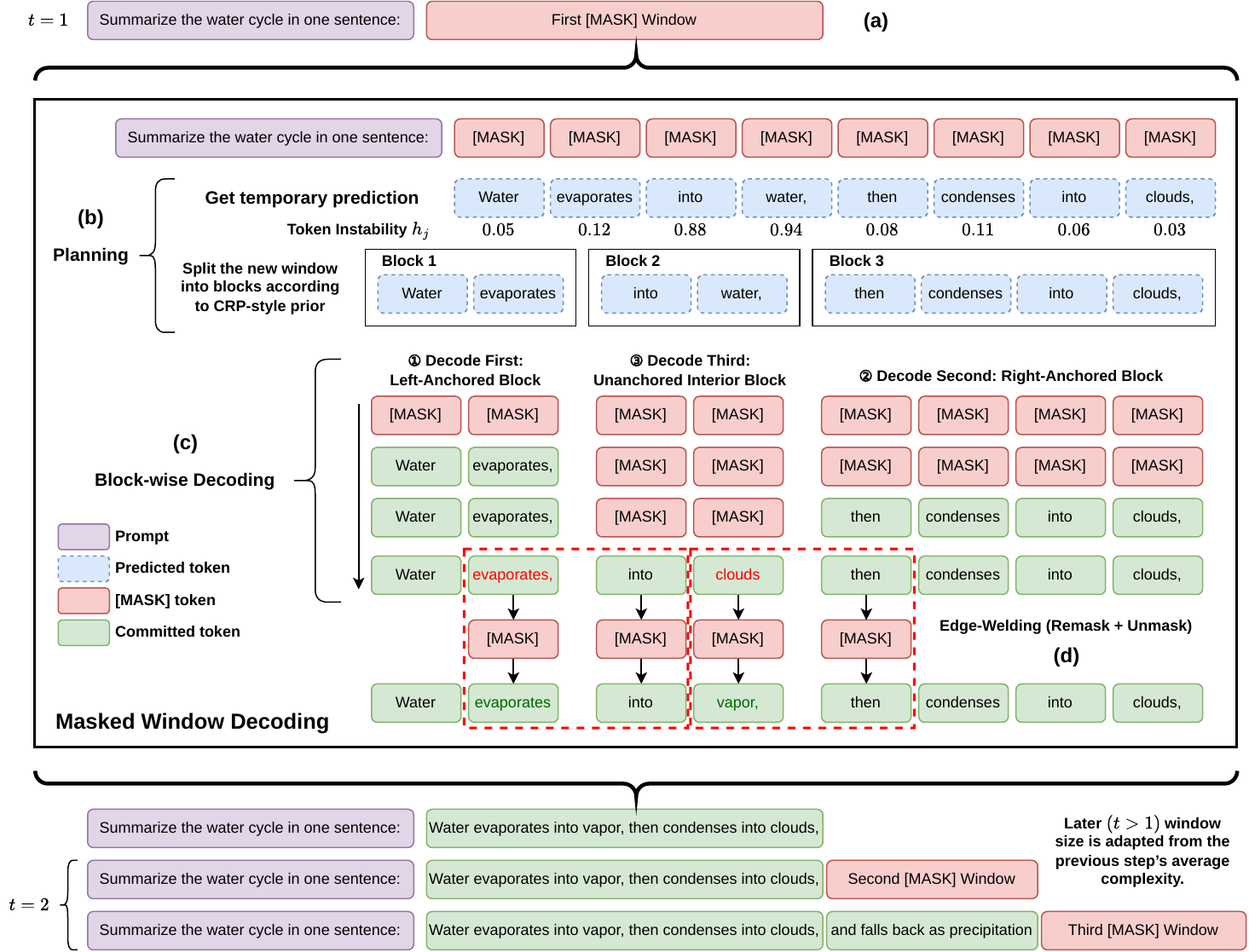}
    \caption{\small \textbf{Overview of \method.} The framework performs flexible-length decoding by iteratively appending masked windows and executing structural inference. \textbf{(a) Window Expansion:} The next window size adaptively scales based on the mean instability ($\bar h$) of previously decoded tokens. \textbf{(b) CRP-Style Partitioning:} A short temporary pass extracts token-level instability scores ($h_j$), which a CRP-style prior uses to partition the unanchored window into contiguous blocks. \textbf{(c) Context-Aware Scheduling:} Partitioned blocks are decoded according to a schedule that prioritizes stable, anchored segments. \textbf{(d) Local Boundary Repair (Edge-Welding):} To align predictive distributions at block interfaces, the decoder performs localized remasking (red dashed boxes) to ensure structural consistency.}
\label{fig:overview}
\end{figure}

\subsection{Dynamic Structured Decoding as Bayesian Inference}

We frame flexible-length diffusion decoding as a dynamic structural inference problem.
We formulate this organization as a unified Bayesian inference over latent variables $Z^{(t)} = \{L_t, \gP^{(t)}, \tau^{(t)}\}$. We denote $O^{(t)}$ as the set of statistics derived from a temporary diagnostic pass that summarizes uncertainty and structural boundary evidence in the current window.
We model the prior over these latent variables as a structured factorization: 
$p\big(Z^{(t)}\big) = p\big(L_t \big)\; p\big(\gP^{(t)} \mid L_t, \alpha\big)\; p\big(\tau^{(t)} \mid \gP^{(t)}\big).$
Here, $p(L_t)$ defines a prior over the window growth size $L_t$. The term $p(\gP^{(t)} \mid L_t, \alpha)$ corresponds to a Chinese Restaurant Process (CRP) prior \citep{blei2010nested} over block partitions $\gP^{(t)}$ across the local indices \(W^{(t)}=\{1,\dots,L_t\}\), where $\alpha$ denotes the concentration parameter. This CRP prior favors coherent contiguous blocks while allowing flexible splitting when supported by edge evidence. Finally, $p(\tau^{(t)} \mid \gP^{(t)})$ defines a preference over the block decoding schedule $\tau^{(t)}$.

Given the prompt $x$, the previously generated sequence $y^{(t-1)}$, and the diagnostic observations $O^{(t)}$, we perform posterior inference over the latent structure:
\begin{equation*}
p\big(Z^{(t)} \mid O^{(t)}, y^{(t-1)}, x\big)
\propto
p\big(O^{(t)} \mid L_t, \gP^{(t)}, \tau^{(t)}, y^{(t-1)}, x\big)\;
p\big(L_t \big)\;
p\big(\gP^{(t)} \mid L_t, \alpha\big)\;
p\big(\tau^{(t)} \mid \gP^{(t)}\big).
\end{equation*}
The structural progression of this inference is depicted in Figure \ref{fig:overview} and summarized in Algorithm \ref{alg:dynamic_decoding_simple}. A detailed algorithm is provided in Appendix \ref{sec:appendix_algo}. To estimate $p\big(Z^{(t)} \mid O^{(t)}, y^{(t-1)}, x\big)$, in the following sections, we will decompose $p\big(Z^{(t)} \mid O^{(t)}, y^{(t-1)}, x\big)$ and present how to estimate the three decomposed posteriors: $p(L_t| O^{(t)}, y^{(t-1)}, x\bigr)$, $p\big(\gP^{(t)} \mid O^{(t)}, L_t, \alpha\big)$, and $p\big(\tau^{(t)} \mid \gP^{(t)}, O^{(t)}\big)$. 

\vspace{-0.2cm}
\begin{algorithm}[ht]
\caption{\method: Dynamic Structured Decoding}
\label{alg:dynamic_decoding_simple}
\vspace{-0.3em}
\begin{algorithmic}[1]
\STATE \textbf{Input:} Prompt $x$, global length limit $N_{\max}$, hyperparameters $(\alpha_0, \gamma, r_{\mathrm{weld}})$
\STATE \textbf{Initialize:} $y^{(0)} \leftarrow x$, step $t \leftarrow 1$, previous window instability $\bar{h}^{(0)} \leftarrow 0.5$
\WHILE{\texttt{[EOS]} not generated \AND $|y^{(t-1)}| < N_{\max}$}
    \STATE Sample expansion length $L_t \sim p(L_t \mid O^{(t)}, y^{(t-1)}, x)$ via Eq. \ref{eq:window_length}
    \STATE Append $L_t$ \texttt{[MASK]} tokens to $y^{(t-1)}$ to form window indices $W^{(t)}$
    \STATE Execute temporary diagnostic pass over $W^{(t)}$ to extract feature signals $\phi^{(t)}$
    \STATE Compute instability $h_j^{(t)}$ (Eq. \ref{eq:stats}) and edge scores $q_g^{(t)}$ for all $j, g \in W^{(t)}$
    \STATE Partition $W^{(t)}$ into blocks $\gP^{(t)}$ using CRP (Eq. \ref{eq:post})
    \STATE Order blocks into schedule $\tau^{(t)}$ via Gibbs distribution (Eq. \ref{eq:schedule})
    \FOR{each block $B \in \tau^{(t)}$}
        \STATE Predict and commit tokens over $T(B)$ refinement steps
    \ENDFOR
    \STATE Perform localized edge-welding at shared block boundaries (Eq. \ref{eq:welding})
    \STATE Calculate finalized mean instability $\bar{h}^{(t)}$ over $W^{(t)}$, update sequence $y^{(t)}$, increment $t$
\ENDWHILE
\STATE \textbf{Output:} Final generated sequence $y^{(\text{final})}$
\end{algorithmic}
\end{algorithm}

\subsection{Latent Block Formation and Growth}

\textbf{Posterior over the new window size.}
At each step $t$, the decoder determines the expansion length $L_t$. A stable preceding window permits a larger window expansion, whereas an unstable window restricts expansion to a smaller window. To capture this dynamic scaling, we summarize the preceding window using its finalized mean instability value \(\bar h^{(t-1)}\in[0,1]\), where larger values indicate structural instability. We model the posterior distribution over the next window length as a Poisson-distributed random variable:
\begin{align} 
\label{eq:window_length} 
L_t \sim p(L_t \mid O^{(t)}, y^{(t-1)}, x\bigr) \approx    \mathrm{Poisson}(\mu_t), \;\;\text{where} \quad \mu_t
=
L_{\min}
+
\bigl(1-\bar h^{(t-1)}\bigr)\bigl(L_{\max}-L_{\min}\bigr),
\end{align}
clipped to \([L_{\min},L_{\max}]\).

\textbf{Statistics for characterizing block boundary changes.}
Before decoding the new window, the model executes a short sequence of temporary diagnostic steps over indices \(W^{(t)}\) to assess positional instability under partial context. At each step, the frozen model predicts all masked positions, commits a fraction of the tokens with the highest confidence, and remasks the remainder. 
For each position \(j \in \{1,\dots,L_t\}\), this diagnostic pass produces a feature vector \(\phi_j^{(t)} \in \mathbb{R}^R\) capturing observable signals including entropy, prediction shifts, hidden state variation, and confidence. This vector is projected to a scalar \(u_j^{(t)}\) using an estimated weight \(w\) (illustrated in Appendix \ref{sec:Appendix_Parameter_Optimization}), and normalized via a logistic function to obtain a local instability score $h_j^{(t)}$:
\begin{equation} \label{eq:stats}
u_j^{(t)} = w^\top \phi_j^{(t)}, \quad
h_j^{(t)} =
\sigma\!\left(
u_j^{(t)} -
\frac{1}{L_t}\sum_{r=1}^{L_t} u_r^{(t)}
\right).
\end{equation}
A larger \(h_j^{(t)}\) indicates higher uncertainty relative to the window. To determine block boundaries, we evaluate the gaps between adjacent tokens. For each gap \(g \in \{1,\dots,L_t-1\}\), we construct a feature vector \(\psi_g^{(t)}\) (illustrated in Appendix~\ref{app:detail}) and project it using an estimated boundary weight vector $w_b$ to compute an edge score:
\begin{equation}
\ell_g^{(t)} = w_b^\top \psi_g^{(t)}, \quad
q_g^{(t)} = \sigma\!\left(\ell_g^{(t)}\right).
\end{equation}
Larger \(q_g^{(t)}\) values indicate stronger probabilistic evidence for placing a boundary at gap \(g\). The window is partitioned using a CRP-inspired prior. We define a local concentration parameter (where $\alpha_0$ is the empirical base constant):
\begin{equation}\label{eq:crp_alpha}
\alpha_g^{(t)} =
\alpha_0
\exp\!\left(
\bar h^{(t-1)} +
\ell_g^{(t)} -
\frac{1}{L_t-1}\sum_{r=1}^{L_t-1}\ell_r^{(t)}
\right).
\end{equation}
Here, $\bar h^{(t-1)}$ controls the overall split rate (higher instability results in more blocks), while $\ell_g^{(t)}$ scores the likelihood of splitting at specific gaps.

\paragraph{Prior over block partitions.}

To model how tokens are grouped into blocks, we place a prior over partitions using the Chinese Restaurant Process (CRP) \citep{blei2010nested}. The key idea is: tokens are assigned sequentially, and each token either joins the current block or starts a new one. Joining the current block is more likely when the block is already large, while starting a new block is controlled by a concentration parameter $\alpha_g^{(t)}$. This naturally balances \emph{block growth} and \emph{block creation}.
At expansion step $t$, the newly allocated window has length $L_t$ and is partitioned into contiguous blocks $\mathcal{P}^{(t)} = \{B_1^{(t)}, \dots, B_{M_t}^{(t)}\}$. We implement the CRP prior through decisions at each of the $L_t - 1$ gaps between adjacent positions. At each gap $g$, the decoder decides whether to continue the current block (``stay'') or start a new block (``cut''). Let $b_g \in \{0,1\}$ denote this decision, with $b_g = 1$ indicating a cut. If the current block has length $m_g$, we define:
\begin{equation}
p(b_g = 0 \mid m_g, \alpha_g^{(t)}) = \frac{m_g}{m_g + \alpha_g^{(t)}}, 
\quad 
p(b_g = 1 \mid m_g, \alpha_g^{(t)}) = \frac{\alpha_g^{(t)}}{m_g + \alpha_g^{(t)}}.
\end{equation}
This formulation has two practical advantages that are important for decoding. First, it \emph{does not require fixing the number of blocks} in advance; the model automatically determines how many blocks are needed. Second, it \emph{does not assume fixed boundaries}; instead, boundaries are inferred dynamically based on local evidence through $\alpha_g^{(t)}$. As a result, the prior encourages coherent block growth (by favoring ``stay'' for large $m_g$) while still allowing new blocks when necessary. The prior probability of a partition $\mathcal{P}^{(t)}$ is then given by:
\begin{equation}
p\big(\mathcal{P}^{(t)} \mid L_t, \alpha_g^{(t)}\big)
=
\prod_{g=1}^{L_t-1}
\left(\frac{\alpha_g^{(t)}}{m_g+\alpha_g^{(t)}}\right)^{b_g}
\left(\frac{m_g}{m_g+\alpha_g^{(t)}}\right)^{1-b_g}.
\end{equation}
\paragraph{Likelihood of diagnostic observations given a block partition.} For each gap $g$, we compute an edge probability $q_g^{(t)} = \sigma(\ell_g^{(t)})$, which we interpret as: $p(b_g = 1 \mid O^{(t)}) = q_g^{(t)}, \quad p(b_g = 0 \mid O^{(t)}) = 1 - q_g^{(t)}$.
 The likelihood of a given partition is:
\begin{equation}
p\big(O^{(t)} \mid \mathcal{P}^{(t)}\big)
=
\prod_{g=1}^{L_t-1}
\big(q_g^{(t)}\big)^{b_g}
\big(1-q_g^{(t)}\big)^{1-b_g}.
\end{equation}
\paragraph{Posterior over block partitions.}
Combining the likelihood and the prior, the posterior over partitions is defined as: $p\big(\mathcal{P}^{(t)} \mid O^{(t)}, L_t, \alpha_g^{(t)}\big)
\propto
p\big(O^{(t)} \mid \mathcal{P}^{(t)}\big)\;
p\big(\mathcal{P}^{(t)} \mid L_t, \alpha_g^{(t)}\big).$

Taking the logarithm, we obtain the objective function:
\begin{equation}
\begin{split}
&\log p\big(\mathcal{P}^{(t)} \mid O^{(t)}, L_t, \alpha_g^{(t)}\big)
=
\sum_{g=1}^{L_t-1}
\Big[
b_g \log q_g^{(t)} 
+ (1-b_g)\log \big(1-q_g^{(t)}\big)
\Big] \\
&\quad +
\sum_{g=1}^{L_t-1}
\Big[
b_g \log \frac{\alpha_g^{(t)}}{m_g+\alpha_g^{(t)}}
+ (1-b_g)\log \frac{m_g}{m_g+\alpha_g^{(t)}}
\Big]
+ A, \;\; \text{where $A$ is a constant}.
\end{split}
\end{equation}
\paragraph{Maximum a posteriori inference determines the block split positions.}
The final partition is obtained via maximum a posteriori (MAP) inference:
\begin{equation}\label{eq:post}
\arg\max_{\mathcal{P}^{(t)}}  \left[  \log p\big(\mathcal{P}^{(t)} \mid O^{(t)}, L_t, \alpha_g^{(t)}\big)\right]
\end{equation}
This objective explicitly grounds the algorithm: the gap feature vectors provide the likelihood evidence for a split, while the CRP prior enforces contiguous block partitions.
Given the resulting split points \(g_1<\dots<g_{M_t-1}\), the contiguous blocks $\mathcal{P}^{(t)}=\{B_1^{(t)},\dots,B_{M_t}^{(t)}\}$ are defined as:
\[
B_1^{(t)}=\{1,\dots,g_1\},\quad
B_2^{(t)}=\{g_1+1,\dots,g_2\},\quad \dots,\quad
B_{M_t}^{(t)}=\{g_{M_t-1}+1,\dots,L_t\}.
\]

\subsection{Blockwise Decoding Schedule and Edge Welding}

\textbf{Posterior over the block decoding schedule.}  
Given a fixed window partition, the decoder assigns each block a refinement budget and a decoding order. For a block \(B \subseteq W^{(t)}\), we define the block instability as $H(B) = \frac{1}{|B|} \sum_{j\in B} h_j^{(t)}$. The total refinement steps \(T(B)\) allocated to the block is obtained by linearly interpolating between \(T_{\min}\) and \(T_{\max}\) using \(H(B)\).

To determine the decoding order, we measure how well a block is anchored by neighboring decoded tokens. Let \(C(B)\in\{0,0.5,1\}\) represent the context proximity: \(C(B)=1\) if both sides are anchored, $0.5$ if one side is anchored, and $0$ if bounded entirely by masks. The schedule \(\tau^{(t)}\) follows a Gibbs distribution:
\begin{equation}\label{eq:schedule}
p\!\left(\tau^{(t)} \mid \mathcal{P}^{(t)}, O^{(t)}, y^{(t-1)}, x\right)
\propto
\exp\!\left(
\sum_{B \in \mathcal{P}^{(t)}} \rho(B)
\right), \quad
\rho(B) = - H(B) + \gamma C(B).
\end{equation}
This schedule prioritizes anchored blocks ($\gamma C(B)$) with low instability ($H(B)$), where $\gamma$ is a context weight. This ordering allows the resulting decoded tokens to serve as stable context that constrains the subsequent decoding of regions with higher instability. Within each scheduled block, the model iteratively commits tokens with high confidence while refining the remaining masked positions.

\textbf{Boundary reconciliation via edge-welding.}  
To ensure distributional consistency across independently scheduled blocks, we apply a local edge-welding step. For neighboring blocks \(B_m^{(t)}=[a,b)\) and \(B_{m+1}^{(t)}=[b,c)\), we define an interval around the boundary:
\begin{equation}\label{eq:welding}
E_m^{(t)} =
\bigl[
\max(a,b-r_{\mathrm{weld}}),
\min(c,b+r_{\mathrm{weld}})
\bigr].
\end{equation}
where $r_{\mathrm{weld}}$ defines the fixed boundary repair radius. Within this interval, tokens with low confidence are remasked and locally refined, while all positions outside the interval remain fixed. This step aligns boundary predictions without modifying the established blocks. After welding is complete, the decoder calculates the updated mean instability \(\bar h^{(t)}\) to control the subsequent expansion step.
\section{Experiments}

\textbf{Baselines and implementation.}   We evaluate \method using LLaDA-8B-Base \citep{nie2025large} and Dream-7B-Base \citep{ye2025dream7bdiffusionlarge}.  To ensure a fair computational comparison, we cap the base unmasking iterations and maximum sequence length at 256. For \method, this iteration limit serves as the budget for dynamic window expansion, whereas baseline models are constrained to generate a rigid 256-token window. We also compare against DAEDAL \citep{li2026beyond}, implementing it as a heuristic plug-in expansion module to represent existing training-free variable-length diffusion methods. \method maintains uniform hyperparameters across all tasks (detailed in Appendix \ref{sec:Appendix_Hyperparameters}). Experiments run on a single NVIDIA H100 GPU.

\textbf{Benchmarks and datasets.} 
To assess generalizability, we benchmark \method across three domains: mathematical reasoning, code synthesis, and symbolic logic. We quantify mathematical reasoning using GSM8K \citep{cobbe2021trainingverifierssolvemath} and MATH \citep{hendrycks2021measuring}, reporting strict-match accuracy. For code generation, we use MBPP \citep{austin2021programsynthesislargelanguage} and HumanEval \citep{chen2021evaluatinglargelanguagemodels}, reporting greedy pass@1 accuracy. Finally, we evaluate multi-step logical reasoning on Big-Bench Hard (BBH) \citep{suzgun-etal-2023-challenging} using exact match accuracy. All evaluations run within the LM-Evaluation-Harness \citep{eval-harness}.

\subsection{Main Results}

As shown in Table \ref{tab:diffusion_performance_final}, \method improves across all five benchmark tests,
among which the score on BBH increases by 4.4 \%
(44.9 $\rightarrow$ 49.3). DAEDAL shows limited improvement on LLaDA-8B-Base and even degrades performance on the BBH task. Besides, to further verify whether this improvement is statistically
reliable at the prompt level, we additionally conduct paired McNemar
tests between \method and DAEDAL on the LLaDA-8B-Base backbone, as
reported in Appendix \ref{sec:appendix_theory_inference}. The paired McNemar tests show that \method achieves statistically
significant prompt-level improvements over DAEDAL on BBH and two Math datasets. For
Code, \method also exhibits a positive prompt-level trend.

Beyond accuracy, this performance gap suggests the instability of DAEDAL when applied to stronger backbone models.
Meanwhile, the results on the more advanced Dream-7B-Base backbone further show
that \method can be effectively transferred beyond LLaDA-8B-Base,
maintaining consistent improvements under a stronger model setting. This
comparison demonstrates the limitations of heuristic rules in
cross-model applications, while highlighting the adaptability of
\method in achieving robust gains through a joint adaptive mechanism of
sequence length control and structured partitioning during the reasoning
stage.
\begin{table*}[ht]
  \centering
  \caption{\small \textbf{Dynamic structured decoding outperforms baselines.} 
DAEDAL serves
as a heuristic length-adjustment decoding baseline. Across these settings,
we keep all shared decoding and evaluation parameters identical wherever
applicable, so that performance differences primarily reflect the effect
of the method-specific decoding mechanisms. Besides, to ensure a fair computational comparison, all models utilize a base budget of 256 unmasking iterations and a maximum generation limit of 256 tokens. \method dynamically scales this base iteration budget across adaptively sized blocks based on block instability. Numbers in parentheses indicate default few-shot examples. Values in gray are standard errors (SE) from \texttt{lm-eval}~\citep{eval-harness}. 
}
  \vspace{-2mm}
  \label{tab:diffusion_performance_final}
  \renewcommand{\arraystretch}{1.18}
  \resizebox{\textwidth}{!}{
    \begin{tabular}{l cc cc c}
      \toprule
      & \multicolumn{2}{c}{\textbf{Code Ability}} & \multicolumn{2}{c}{\textbf{Mathematics}} & \textbf{General} \\
      \cmidrule(lr){2-3} \cmidrule(lr){4-5} \cmidrule(l){6-6}
      \textbf{Model} & HumanEval (0) & MBPP (3) & GSM8K (5) & MATH (4) & BBH (3) \\
      \midrule
      LLaDA-8B-Base & 32.3\stderr{3.6} & 39.8\stderr{2.1} & 70.3\stderr{1.2} & 30.5\stderr{0.6} & 44.9\stderr{0.5} \\
      LLaDA-8B-Base + DAEDAL & 33.5\stderr{3.7} & 40.2\stderr{2.1} & 70.8\stderr{1.2} & 31.2\stderr{0.6} & 43.7\stderr{0.5} \\
      \textbf{LLaDA-8B-Base + \method} & \textbf{34.8}\stderr{3.7} & \textbf{41.4}\stderr{2.2} & \textbf{72.1}\stderr{1.2} & \textbf{31.4}\stderr{0.6} & \textbf{49.3}\stderr{0.5} \\
      \midrule
      Dream-7B-Base & 40.2\stderr{2.7} & 57.2\stderr{2.2} & 74.9\stderr{1.1} & 38.2\stderr{0.6} & 51.7\stderr{0.5} \\
      Dream-7B-Base + DAEDAL & 34.7\stderr{2.7} & 54.4\stderr{2.2} & 74.3\stderr{1.2} & 38.6\stderr{0.6} & 44.8\stderr{0.5} \\
      \textbf{Dream-7B-Base + \method} & \textbf{47.0}\stderr{3.9} & \textbf{59.8}\stderr{2.1} & \textbf{75.1}\stderr{1.9} & \textbf{38.8}\stderr{0.6} & \textbf{52.5}\stderr{0.5} \\
      \bottomrule
    \end{tabular}
  }
\end{table*}

\begin{wrapfigure}{t!}{0.5\columnwidth} 
  \centering
\includegraphics[width=0.8\linewidth]{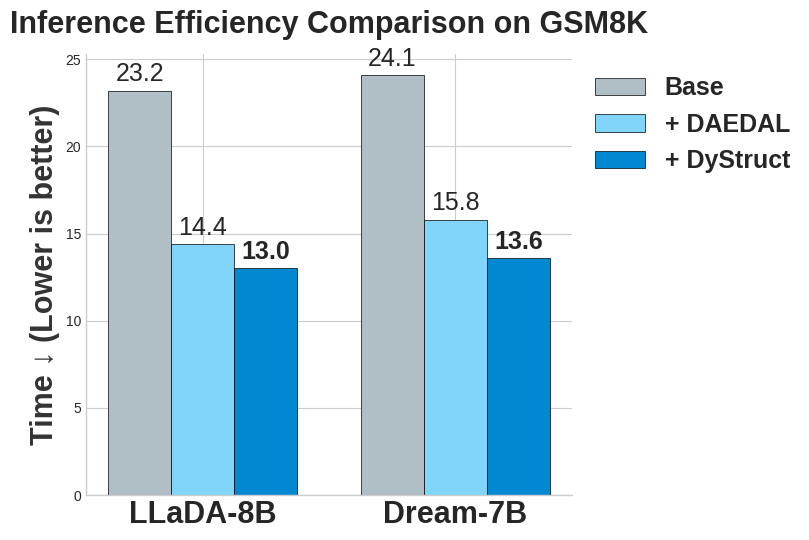} 
  \caption{\small \textbf{Inference efficiency comparison.} The proposed \method achieves the lowest inference time across different backbone models on the GSM8K dataset. Time is reported in seconds per iteration (s/it).}
\label{fig:efficiency_comparison}
\end{wrapfigure}

\textbf{Dynamic structure and instability-aware scheduling improve inference performance.}
Beyond the aggregate accuracy gains, the results also
suggest that \method improves efficiency by allocating computation
according to structural instability rather than applying a uniform
decoding budget. This is particularly reflected on BBH, the most
reasoning-intensive benchmark in our evaluation, where \method
achieves the largest improvement. Since BBH contains more complex
multi-step reasoning and logical composition, this gain is consistent
with the role of instability-aware scheduling: high-difficulty regions,
such as logical operators, intermediate reasoning steps, and structured
expressions, receive more refinement, while easier regions are decoded
with fewer redundant updates.
Figure \ref{fig:efficiency_comparison} shows that this adaptive allocation also leads to faster
inference on GSM8K. Because GSM8K mainly consists of structured mathematical reasoning problems, many output regions
become stable early once the key arithmetic or reasoning pattern is
resolved.
\method can therefore avoid repeatedly denoising the entire
fixed-length window and instead spend computation only where additional
refinement is needed. This explains why \method achieves lower
per-question inference time than both the base decoding setup and DAEDAL
on LLaDA-8B and Dream-7B. 
These results show that \method
does not merely improve accuracy on difficult reasoning tasks such as
BBH, but also reduces unnecessary computation on mathematical
tasks by dynamically matching the decoding budget to block instability.

\begin{table*}[bt]
  \centering
  \caption{\small \textbf{Ablation study of structural decoding components.}
  Both the Bayesian block partitioning and the local boundary repair step are necessary to maintain reasoning accuracy.
  W/o  Block Decoding Schedule replaces the context-aware block scheduling with a fixed left-to-right block order.}
  \vspace{-2mm}
\label{tab:ablation_components}
\renewcommand{\arraystretch}{1.1}
  \resizebox{\textwidth}{!}{
    \begin{tabular}{l cc cc c}
      \toprule
      & \multicolumn{2}{c}{\textbf{Coding Ability}} 
      & \multicolumn{2}{c}{\textbf{Mathematics}} 
      & \textbf{General} \\
      \cmidrule(lr){2-3}
      \cmidrule(lr){4-5}
      \cmidrule(l){6-6}
      \textbf{Configuration} & HumanEval & MBPP & GSM8K & MATH & BBH \\
      \midrule
      \rowcolor{mainblue}
      Full \method 
      & \textbf{34.8} 
      & \textbf{41.4} 
      & \textbf{72.1} 
      & \textbf{31.4} 
      & \textbf{49.3} \\ 
      
      w/o Block Decoding Schedule          
      & 33.5 & 41.2 & 71.8 & 30.3 & 49.3 \\
      
      w/o Boundary Repair          
      & 32.9 & 40.8 & 70.5 & 31.2 & 49.1 \\
      
      w/o Block Decoding Schedule \& Boundary Repair  
      & 32.9 & 41.0 & 69.7 & 31.0 & 49.1 \\

      \midrule
      LLaDA-8B-Base 
      & 32.3 & 40.2 & 70.3 & 30.5 & 44.9 \\
      \bottomrule
    \end{tabular}
  }
  \vspace{-2mm}
\end{table*} 

\begin{table*}[t]
  \centering
  \setlength{\tabcolsep}{3.5pt}
  \renewcommand{\arraystretch}{0.83}
  \begin{minipage}{0.49\textwidth}
    \centering
    \caption{\small \textbf{Effect of concentration parameter.}
    The default setting is shaded in gray.
    \textit{Toks} denotes the average number of newly generated tokens per sample, and \textit{Blks} denotes the rounded average number of finalized blocks per sample. We vary the concentration parameter $\alpha_0$ and the results remain stable across different $\alpha_0$.}
    \label{tab:sensitivity_alpha}
    {\small
    \begin{tabular}{@{}lccc@{\hspace{0.9em}}ccc@{}}
      \toprule
      & \multicolumn{3}{c}{\textbf{HumanEval}}
      & \multicolumn{3}{c}{\textbf{BBH}} \\
      \cmidrule(lr){2-4}
      \cmidrule(lr){5-7}
      \boldmath$\alpha_0$ 
      & \textbf{Pass@1}$\uparrow$ & \textbf{Toks}$\downarrow$ & \textbf{Blks}
      & \textbf{Acc.}$\uparrow$ & \textbf{Toks}$\downarrow$ & \textbf{Blks} \\
      \midrule

      1.2 & 35.4 & 217 & 7
          & 49.4 & 246 & 7 \\
      \rowcolor{black!6}
      1.5 & 34.8 & 219 & 7
          & 49.3 & 246 & 7 \\

      1.8 & 33.5 & 220 & 7
          & 49.5 & 245 & 6 \\

      2.0 & 34.1 & 221 & 6
          & 49.3 & 245 & 6 \\
      \bottomrule
    \end{tabular}
    }
  \end{minipage}
  \hfill
  \begin{minipage}{0.49\textwidth}
    \centering

    \caption{\small \textbf{Effect of initial sequence length.}
    We vary the initial sequence length to evaluate the stability of our method. Performance remains relatively stable across different initial sequence length. The default setting is shaded in gray.}
    \label{tab:sensitivity_length}
    {\small
    \begin{tabular}{@{}lccc@{\hspace{0.9em}}ccc@{}}
      \toprule
      & \multicolumn{3}{c}{\textbf{HumanEval}} 
      & \multicolumn{3}{c}{\textbf{BBH}} \\
      \cmidrule(lr){2-4}
      \cmidrule(lr){5-7}
      \textbf{Len.} 
      & \textbf{Pass@1}$\uparrow$ & \textbf{Toks}$\downarrow$ & \textbf{Blks}
      & \textbf{Acc.}$\uparrow$ & \textbf{Toks}$\downarrow$ & \textbf{Blks} \\
      \midrule
      32  & 33.5 & 220 & 8
          & 48.8 & 239 & 7 \\

      \rowcolor{black!6}
      48  & 34.8 & 219 & 7
          & 49.3 & 246 & 7 \\

      64  & 33.5 & 216 & 6
          & 49.0 & 246 & 6 \\

      128 & 35.4 & 225 & 5
          & 48.9 & 248 & 5 \\

      256 & 36.6 & 256 & 7
          & 46.3 & 256 & 3 \\
      \bottomrule
    \end{tabular}
    }
  \end{minipage}
\end{table*}

\textbf{Structural components drive stable decoding performance.} Table \ref{tab:ablation_components} isolates the contribution of each structural mechanism. Removing block decoding leads to lower mathematical reasoning accuracy on both benchmarks, as the procedure loses the benefit of dynamically
prioritizing structurally stable blocks during decoding. Furthermore, removing
the local boundary repair step degrades code generation performance, reducing HumanEval accuracy by 1.9 points. This performance degradation
demonstrates that independently scheduled blocks require localized edge-welding
to align their predictive distributions; without it, adjacent blocks remain
structurally incompatible. When both the block decoding schedule and
boundary repair are removed, the degradation becomes more apparent, with GSM8K
further decreasing to 69.7\% and MATH to 31.0\%. Notably, even after removing
these components, the remaining CRP-based block partitioning mechanism still
outperforms the LLaDA-8B-Base baseline on most benchmarks and maintains strong
performance on BBH. This indicates that the core structural partitioning itself
already provides a robust benefit, while the block decoding schedule
and boundary repair further improve decoding stability and task performance.

\subsection{Ablation, Sensitivity and Qualitative Analysis}

\textbf{Adaptive window expansion mitigates fragmentation without sacrificing accuracy.} 
As demonstrated in Table \ref{tab:sensitivity_length}, \method remains robust across different
initial window lengths. Although HumanEval reaches its highest Pass@1
under the full 256-token setting, the selected length of 48 provides
a stronger overall balance, maintaining competitive HumanEval performance
while achieving the best BBH accuracy. Table \ref{tab:sensitivity_alpha} further supports the mechanism behind this behavior. Across
different $\alpha_0$ and initial length settings, \method maintains
stable performance while using a compact effective token budget and a
small number of blocks in most configurations, rather than
relying on uniformly expanding the full 256-token window. This indicates
that the model can adapt the active decoding region according to the
local structure of the output. By partitioning the window into contiguous blocks and scheduling
computation based on local instability $h_j$, \method decouples generation
length from the global computational budget. Instead of assigning the
same number of unmasking iterations to the entire sequence, the total
iterations for each block are interpolated from a base iteration count
and the mean instability score $H(B)$. As a result, high-instability segments
receive near-maximum iteration counts, whereas stable linguistic
transitions and low-complexity regions receive fewer unmasking
iterations. This explains why \method can improve difficult reasoning
tasks such as BBH while also reducing redundant computation on simpler
regions.

\textbf{Hyperparameters balance granularity and context.}
The effect of the CRP concentration prior $\alpha_0$ has been analyzed
above in Table \ref{tab:sensitivity_alpha}, where \method shows stable performance under
variations of the partition granularity. Beyond $\alpha_0$, \method
relies on several architectural hyperparameters to govern block
resolution. The welding radius
$r_{\mathrm{weld}}$ defines the local interval used for boundary repair. We provide additional ablations of these architectural hyperparameters,
including the welding radius $r_{\mathrm{weld}}$, in Appendix \ref{sec:Ablative_Studies}. These
results show that \method remains stable under variations of
their values; only overly small or overly large settings lead to slight
performance changes. Thus, we adopt fixed values that provide stable partitioning across all datasets rather than tuning them for
task-specific optimality, demonstrating the robustness of the structural
formulation without heavy hyperparameter sweeping.

\textbf{Qualitative Results}
To illustrate how \method treats flexible-length generation as a problem of dynamic structure, we provide examples in Appendix \ref{sec:qualitative}. Cases demonstrate the execution of CRP-inspired block partitioning, context-aware scheduling, and localized edge-welding. While baseline models frequently commit to context-brittle tokens early in the sequence and propagate structural failure forward, \method utilizes local boundary repair to identify and revise conflicting boundaries as the context evolves, recovering structurally valid outputs on complex mathematical prompts.

\textbf{Efficiency Analysis}
As shown in Figure \ref{fig:efficiency_comparison} , \method reduces inference latency on GSM8K from 23.2s to 13.0s for LLaDA-8B-Base and from 24.1s to 13.6s for Dream-7B-Base, outperforming both standard decoding and DAEDAL. Although DyStruct introduces slightly more model forward passes, its adaptive window expansion starts with 48 tokens and progressively extends the sequence, avoiding repeated computation over a fixed 256-token window. As shown in Table \ref{tab:sensitivity_alpha}, the average realized generation lengths are approximately 219 tokens on HumanEval and 246 on BBH. Thus, the additional forward passes process shorter sequences on average and therefore are cheaper in terms of FLOPs.
\section{Conclusion}

This paper presents a principled Bayesian framework for flexible-length diffusion language models (DLMs). We formulate flexible-length generation as a joint posterior inference problem over dynamic window expansion, latent block structure, and decoding organization. Extensive experiments across multiple benchmarks show that our approach consistently outperforms both fixed-length and existing flexible-length DLMs decoding methods on a variety of datasets.

\textbf{Limitations} Our method operates purely at inference time without modifying model parameters. While this enables broad applicability, integrating structural inference into training may further enhance performance, which we leave for future work.

\textbf{Impact Statements}    This paper presents work whose goal is to advance the field of machine learning. There are many potential societal consequences of our work, none of which we feel must be specifically highlighted here.

\section{Acknowledgments}
We acknowledge the University of Central Florida Advanced Research Computing Center (ARCC) for providing computational resources, as well as the Advanced Cyberinfrastructure Coordination Ecosystem: Services \& Support (ACCESS) program for computational resources.

\bibliographystyle{plainnat}
\bibliography{reference}

\appendix
\label{appendix}
\newpage
\section{Notation}
\label{sec:appendix_Notation}

Table \ref{tab:notation} provides a concise mathematical reference for the notation used in our method.

\begin{table}[h]
  \centering
  \caption{Mathematical notation for \method.}
  \label{tab:notation}
  \renewcommand{\arraystretch}{1.2}
  \begin{tabular}{ll}
    \toprule
    Symbol & Description \\
    \midrule
    \multicolumn{2}{l}{\textit{Decoding and Sequence Parameters}} \\
    \midrule
    $t$ & Window expansion step index. \\
    $L_t$ & Window expansion length at step $t$. \\
    $L_{\min}, L_{\max}$ & Bounds for the window expansion length. \\
    $N_{\max}$ & Global sequence token limit. \\
    $W^{(t)}$ & Local indices in the newly appended window. \\
    $Z^{(t)}$ & Set of latent structural variables $\{L_t, \mathcal{P}^{(t)}, \tau^{(t)}\}$. \\
    \midrule
    \multicolumn{2}{l}{\textit{Diagnostic Temporary Pass and Uncertainty Metrics}} \\
    \midrule
    $O^{(t)}$ & Observations from the diagnostic temporary pass. \\
    $p_g^{\mathrm{temp}}$ & Initial distribution from the diagnostic pass. \\
    $\mathcal{H}_i$ & Token predictive entropy at position $i$. \\
    $\phi_j^{(t)}$ & Local feature vector at token position $j$. \\
    $w$ & Learned weights for position features $\phi_j^{(t)}$. \\
    $h_j^{(t)}$ & Position instability score at position $j$. \\
    $\bar{h}^{(t)}$ & Mean instability of the previously decoded window. \\
    \midrule
    \multicolumn{2}{l}{\textit{Partitioning and CRP Prior}} \\
    \midrule
    $g$ & Gap index between token positions $g$ and $g+1$. \\
    $\psi_g^{(t)}$ & Gap feature vector for the boundary at gap $g$. \\
    $w_b$ & Learned weights for gap features $\psi_g^{(t)}$. \\
    $\ell_g^{(t)}, q_g^{(t)}$ & Logit edge score and boundary split probability. \\
    $\alpha_g^{(t)}$ & Local CRP concentration parameter at gap $g$. \\
    $\alpha_0$ & Base CRP concentration parameter. \\
    $m_g$ & Length of the block immediately preceding gap $g$. \\
    $\mathcal{P}^{(t)}$ & Partition of $W^{(t)}$ into contiguous token blocks. \\
    $B_m^{(t)}$ & The $m$-th block in partition $\mathcal{P}^{(t)}$. \\
    \midrule
    \multicolumn{2}{l}{\textit{Scheduling and Edge-Welding}} \\
    \midrule
    $\tau^{(t)}$ & Context-aware decoding schedule for partition $\mathcal{P}^{(t)}$. \\
    $H(B)$ & Mean instability score of block $B$. \\
    $C(B)$ & Context-adjacency metric for block $B$. \\
    $\gamma$ & Weight for context-adjacency priority. \\
    $T(B)$ & Interpolated refinement steps allocated to block $B$. \\
    $T_{\min}, T_{\max}$ & Bounds for block refinement steps. \\
    $r_{\mathrm{weld}}$ & Spatial radius for localized edge-welding. \\
    $E_m^{(t)}$ & Localized edge-welding interval between adjacent blocks. \\
    \bottomrule
  \end{tabular}
\end{table}
\newpage
\section{Algorithm}
\label{sec:appendix_algo}

Algorithm \ref{alg:dynamic_decoding} details the execution \method. The process transforms flexible-length generation into an iterative cycle of adaptive expansion, structural partitioning, and context-aware resolution.

\begin{algorithm}[ht]
\caption{\method: Dynamic Structured Decoding}
\label{alg:dynamic_decoding}
\begin{algorithmic}[1]
\STATE \textbf{Input:} Prompt $x$, global length limit $N_{\max}$, hyperparameters $(\alpha_0, \gamma, r_{\mathrm{weld}})$
\STATE \textbf{Initialize:} $y^{(0)} \leftarrow x$, \quad step $t \leftarrow 1$, \quad previous window instability $\bar{h}^{(0)} \leftarrow 0.5$
\WHILE{\texttt{[EOS]} not generated \AND $|y^{(t-1)}| < N_{\max}$}
    
    \vspace{0.15cm}
    \STATE \textit{\# 1. Adaptive Window Expansion}
    \STATE Sample expansion length $L_t \sim p(L_t \mid O^{(t)}, y^{(t-1)}, x)$ via $\mathrm{Poisson}(\mu_t)$ scaled by $(1 - \bar{h}^{(t-1)})$
    \STATE Append $L_t$ \texttt{[MASK]} tokens to $y^{(t-1)}$ to form local window indices $W^{(t)}$
    
    \vspace{0.15cm}
    \STATE \textit{\# 2. Diagnostic Pass \& Partitioning}
    \STATE Execute temporary diagnostic pass over $W^{(t)}$ to extract feature signals $\phi^{(t)}$
    \STATE Compute position instability $h_j^{(t)}$ and gap split probabilities $q_g^{(t)}$ for all $j, g \in W^{(t)}$
    \STATE Evaluate local CRP parameters $\alpha_g^{(t)}$ against continuation and cutting scores
    \STATE Partition $W^{(t)}$ into contiguous blocks $\mathcal{P}^{(t)} = \{B_1^{(t)}, \dots, B_{M_t}^{(t)}\}$
    
    \vspace{0.15cm}
    \STATE \textit{\# 3. Instability-Aware Scheduling}
    \STATE Compute mean instability $H(B)$ and context-adjacency $C(B)$ for all $B \in \mathcal{P}^{(t)}$
    \STATE Order blocks into schedule $\tau^{(t)}$ by sorting $-H(B) + \gamma C(B)$
    
    \vspace{0.15cm}
    \STATE \textit{\# 4. Block-wise Resolution \& Edge-Welding}
    \FOR{each block $B \in \tau^{(t)}$}
        \STATE Predict and commit tokens over interpolated $T(B)$ refinement steps
    \ENDFOR
    \FOR{each shared boundary between adjacent blocks in $\mathcal{P}^{(t)}$}
        \STATE Remask lowest-confidence positions strictly inside radius $r_{\mathrm{weld}}$ and refine
    \ENDFOR
    
    \vspace{0.15cm}
    \STATE \textit{\# 5. State Update}
    \STATE Calculate finalized mean instability $\bar{h}^{(t)}$ over the decoded window $W^{(t)}$
    \STATE Update sequence $y^{(t)}$ and increment $t \leftarrow t + 1$
\ENDWHILE
\STATE \textbf{Output:} Final generated sequence $y^{(\text{final})}$
\end{algorithmic}
\end{algorithm}

\paragraph{Algorithmic Description.} 
At each window expansion step, the decoder determines the length of the new masked window by evaluating the stability of the previously generated segment; high instability restricts expansion to prevent the propagation of structural errors. Before unmasking begins, a temporary diagnostic pass extracts local feature signals $\phi^{(t)}$. These diagnostic features inform a Bayesian partitioning step where a CRP prior groups tokens into contiguous blocks based on local instability. To ensure stable conditioning, the decoder resolves these partitioned blocks in a context-aware order, prioritizing segments that exhibit low-instability or exist adjacent to established context. Finally, an edge-welding step performs localized remasking at the interfaces of adjacent blocks to reconcile the predictive distributions and ensure sequence coherence.

\section{Method Details}
\label{app:detail}

We form a gap feature vector:
\begin{equation}
\psi_g^{(t)} = \left[h_g^{(t)}, h_{g+1}^{(t)}, \left|h_g^{(t)} - h_{g+1}^{(t)}\right|, \mathrm{JSD}(p_g^{\mathrm{temp}} \parallel p_{g+1}^{\mathrm{temp}})\right],
\label{eq:gap_features}
\end{equation}
where $p_g^{\mathrm{temp}}$ is the initial diagnosis distribution at position $g$, and $\mathrm{JSD}(\cdot \parallel \cdot)$ represents the Jensen-Shannon divergence. This gap feature vector quantifies whether the positions bounding the gap operate as a single block or require structural separation.

\section{Calibration of Instability Coefficients}
\label{sec:Appendix_Parameter_Optimization}

To determine the instability coefficients ($w$), we construct a calibration dataset $\mathcal{D} = \{(\phi_i, d_i)\}_{i=1}^N$ by extracting $N$ token-level observations from validation trajectories with accessible ground-truth sequences. We define the feature vector $\phi_i \in \mathbb{R}^7$ such that each vector contains the diagnostic metrics recorded for a specific token:
\begin{equation}
\phi_i = [\mathcal{H}_i, R_i, \Omega_i, \mathrm{JSD}_i, \Delta s_i, F_i, G_i]^\top
\label{eq:feature_vector}
\end{equation}
where the features correspond to predictive entropy ($\mathcal{H}$), remasking frequency ($R$), logit oscillation ($\Omega$), Jensen-Shannon divergence ($\mathrm{JSD}$), hidden state jump ($\Delta s$), confidence ($F$), and probability margin ($G$). We define the binary diagnostic targets $d_i \in \{0, 1\}$ using the ground-truth sequences: $d_i = 1$ if the predicted token mismatches the ground truth or triggers a remasking event during decoding. Conversely, $d_i = 0$ if the token correctly matches the ground truth and remains committed.

The components of $\phi_i$ are defined as follows:
\begin{subequations}\label{eq:diagnostic_features}
\begin{align}
\mathcal{H}_i &=
-\sum_{v\in\mathcal{V}} p_{i,v}\log p_{i,v},\\
R_i &=
\frac{1}{K}\sum_{k=1}^{K}
\mathbf{1}\!\left[\kappa_i^{(k)}=1 \land \eta_i^{(k)}=0\right],\\
\Omega_i &=
\frac{1}{\max(K-1,1)}
\sum_{k=2}^{K}
\mathbf{1}\!\left[
\hat{y}_i^{(k)}\neq \hat{y}_i^{(k-1)}
\right],\\
\mathrm{JSD}_i &=
\frac{1}{\max(K-1,1)}
\sum_{k=2}^{K}
\mathrm{JSD}\!\left(p_i^{(k)}\parallel p_i^{(k-1)}\right),\\
\Delta s_i &=
\frac{1}{d}\sum_{r=1}^{d}
\left|s_{i,r}-s_{i-1,r}\right|,\\
F_i &= p_{i,\hat{y}_i},\\
G_i &= \log p_{i,(1)}-\log p_{i,(2)}.
\end{align}
\end{subequations}
Here $z_i \in \mathbb{R}^{|\mathcal{V}|}$ denotes the model logits at position $i$, $s_i \in \mathbb{R}^d$ denotes the corresponding final-layer hidden state, $p_i=\mathrm{softmax}(z_i)$, $\hat{y}_i=\arg\max_v z_{i,v}$, and $p_{i,(1)}$ and $p_{i,(2)}$ denote the largest and second-largest probabilities at position $i$, respectively. The boolean variable $\kappa_i^{(k)}$ indicates if the position is masked at refinement step $k$, and $\eta_i^{(k)}$ indicates if the token is accepted. The Jensen--Shannon divergence is defined as:
\begin{equation}
\mathrm{JSD}(p\parallel q)
=
\frac{1}{2}\mathrm{KL}(p\parallel m)
+
\frac{1}{2}\mathrm{KL}(q\parallel m),
\qquad
m=\frac{1}{2}(p+q).
\label{eq:jsd_definition}
\end{equation}

We estimate the optimal coefficient vector $w^*$ by minimizing an $L_2$-regularized binary cross-entropy loss over the calibration dataset $\mathcal{D}$:
\begin{equation}
\mathcal{J}(w) = -\frac{1}{N} \sum_{i=1}^N \left[ d_i \log \sigma(w^\top \phi_i) + (1 - d_i) \log(1 - \sigma(w^\top \phi_i)) \right] + \lambda_{\mathrm{reg}} \|w\|_2^2
\label{eq:bce_loss}
\end{equation}
where $\sigma(\cdot)$ is the sigmoid function and $\lambda_{\mathrm{reg}}$ is the regularization penalty. 

During inference, the runtime algorithm evaluates the unconstrained logit $u_i$:
\begin{equation}
u_i = (w^*)^\top \phi_i
\label{eq:linear_logit}
\end{equation}
The linear projection $u_i$ preserves the relative ranking of token instability established during calibration. The generation pipeline then normalizes the projection $u_i$ using the window-centered logistic function defined in Equation \ref{eq:stats} to compute the final positional instability score $h_i$.

\section{Implementation Details and Hyperparameters}
\label{sec:Appendix_Hyperparameters}

To ensure the generalizability of \method,
\method maintains a uniform set of hyperparameters across all benchmarks, domains, and model scales. We do not tune these parameters for task-specific optimality.

\textbf{Window Expansion.} The dynamic expansion loop is bounded by a minimum burst length of $L_{\min} = 8$ and a maximum burst length of $L_{\max} = 48$. 

\textbf{Partitioning and Scheduling.} The Bayesian partitioning utilizes an empirical base CRP concentration prior of $\alpha_0 = 1.5$. During schedule evaluation, the context-adjacency priority weight is set to $\gamma = 2.0$.

\textbf{Blockwise Decoding and Welding.} Based on the block instability score $H(B) \in [0, 1]$, the total refinement steps $T(B)$ are interpolated between $T_{\min} = 6$ and $T_{\max} = 18$. Finally, the localized edge-welding step operates within a fixed spatial repair radius of $r_{\mathrm{weld}} = 4$ tokens and executes for $4$ refinement steps.
\section{Qualitative Examples}
\label{sec:qualitative}

We provide decoding traces for code synthesis, logical reasoning, and linguistic disambiguation to demonstrate the mechanical execution of CRP-inspired partitioning, context-aware scheduling, and localized edge-welding. 

\paragraph{Localized edge-welding resolves boundary fragmentation through targeted re-evaluation.} 
Baseline diffusion models frequently produce distributional shifts at boundaries during flexible-length generation. Figure \ref{fig:qualitative-code-example} illustrates this shift in a \textbf{HumanEval} programming task. When Block 2 is resolved independently, the partial context \texttt{if abs} provides an insufficient conditional signal, resulting in incoherent variables (\texttt{x - y}). \method utilizes localized edge-welding to target the predictive entropy ($\mathcal{H}$) spike at this specific interface. By forcing a targeted re-evaluation strictly within the boundary overlap, the framework reconciles the adjacent distributions to recover context-grounded syntax \texttt{(numbers[i] - numbers[j])}.

\definecolor{framegray}{gray}{0.8}
\begin{figure}[H]
\centering
\begin{tcolorbox}[width=\columnwidth,
  colback=white,colframe=framegray,boxrule=0.6pt,arc=2mm,
  left=4pt,right=4pt,top=4pt,bottom=4pt]
\small
\setlength{\tabcolsep}{6pt}
\renewcommand{\arraystretch}{1.2}
\setlength{\fboxsep}{1pt}
\begin{tabular}{@{}p{0.25\linewidth} p{0.70\linewidth}@{}}
\textbf{Dataset:} & \texttt{HumanEval} \\
\textbf{Prompt:} & \texttt{Given an array \texttt{numbers}, determine if any two elements have an absolute difference less than a threshold.} \\
\midrule
\textbf{Context:} & \texttt{def check\_diff(numbers, threshold):} \\
                  & \texttt{~~~~for i in range(len(numbers)):} \\
\midrule
\textbf{Pre-Welding:} & \texttt{for j in range(i + 1, len(numbers)):} \\
\textit{(Independent Blocks)} & \texttt{~~~~if abs\colorbox{red!25}{\strut (x - y)} < threshold:} \\
\midrule
\textbf{Post-Welding:} & \texttt{for j in range(i + 1, len(numbers)):} \\
\textit{(Final Output)} & \texttt{~~~~if abs\colorbox{green!25}{\strut (numbers[i] - numbers[j])} < threshold:} \\
\end{tabular}
\end{tcolorbox}
\caption{\textbf{\method Resolves Boundary Fragmentation via Edge-Welding.} Independent block decoding produces structurally incompatible boundaries. In this trace, Block 2 produces incoherent variables (\texttt{x - y}) due to weak boundary conditioning. \method targets the entropy spike at this interface, replacing the incoherent segment with context-grounded syntax. (\textcolor{red}{Red}: incoherent variables; \textcolor{green}{Green}: localized repair.)}
\label{fig:qualitative-code-example}
\end{figure}

\paragraph{CRP-inspired partitioning isolates high-instability regions to constrain error propagation.}
Beyond coordinating boundaries, \method adapts the internal structure of the generated region based on local instability. Figure \ref{fig:qualitative-bbh-logic} demonstrates the framework partitioning a window during a \textbf{Big-Bench Hard (BBH)} evaluation. The CRP prior evaluates the instability signals ($h_j^{(t)}$) from the temporary diagnostic pass and partitions the sequence into two distinct blocks. The context-aware schedule identifies the initial segment as well-anchored and low-instability ($H(B)\!=\!0.27$), prioritizing its resolution. Prioritizing stable segments first establishes a reliable conditional anchor before the decoder refines the high-instability deductive segment in Block 2 ($H(B)\!=\!0.52$).

\begin{figure}[H]
\centering
\begin{tcolorbox}[width=\columnwidth,
  colback=white,colframe=framegray,boxrule=0.6pt,arc=2mm,
  left=4pt,right=4pt,top=4pt,bottom=4pt]
\small
\setlength{\tabcolsep}{6pt}
\renewcommand{\arraystretch}{1.2}
\setlength{\fboxsep}{1pt}
\begin{tabular}{@{}p{0.25\linewidth} p{0.70\linewidth}@{}}
\textbf{Dataset:} & \texttt{BBH Logical Deduction} \\
\textbf{Prompt:} & \texttt{Evaluate the logical validity: Let A = True. B = (not A). Is A and B True?} \\
\midrule
\textbf{Context:} & \texttt{Let's evaluate B: B = ( True ) = True. Plugging} \\
\midrule
\textbf{Block 1 (Low $\mathcal{H}$):} & \texttt{\colorbox{blue!15}{\strut in A and B,}} \\
\textit{Schedule Priority 1} & \textit{Length: 5 tokens | Mean Instability: 0.27 | Steps: 41} \\
\midrule
\textbf{Block 2 (High $\mathcal{H}$):} & \texttt{\colorbox{red!15}{\strut we get True and False, which evaluates to False.}} \\
\textit{Schedule Priority 2} & \textit{Length: 12 tokens | Mean Instability: 0.52 | Steps: 215} \\
\end{tabular}
\end{tcolorbox}
\caption{\textbf{\method Isolates Logical Transitions via Partitioning.} The framework splits the unanchored window to isolate segments with high instability scores. The decoding schedule prioritizes the low-instability segment (Block 1) to provide stable conditioning before refining the logical evaluation in Block 2. (\textcolor{blue}{Blue}: low-instability segment; \textcolor{red}{Red}: high-instability deduction.)}
\label{fig:qualitative-bbh-logic}
\end{figure}

\paragraph{Context-aware scheduling utilizes stable anchors to provide bidirectional conditioning.}
In tasks containing causal dependencies, the decoding order determines the stability of the generated context. Figure \ref{fig:qualitative-bbh-disambiguation} details a three-block trace from a \textbf{BBH Disambiguation} task. By identifying both the initial setup and the final conclusive answer as stable segments ($H\!=\!0.21$ and $H\!=\!0.18$), the scheduler resolves these terminal blocks first. This terminal-block-first scheduling strategy provides bidirectional context to ground the high-instability inferential segment in Block 2, ensuring the pronoun resolution is consistent with both the preceding setup and the required output format.

\begin{figure}[H]
\centering
\begin{tcolorbox}[width=\columnwidth,
  colback=white,colframe=framegray,boxrule=0.6pt,arc=2mm,
  left=4pt,right=4pt,top=4pt,bottom=4pt]
\small
\setlength{\tabcolsep}{6pt}
\renewcommand{\arraystretch}{1.2}
\setlength{\fboxsep}{1pt}
\begin{tabular}{@{}p{0.25\linewidth} p{0.70\linewidth}@{}}
\textbf{Dataset:} & \texttt{BBH Disambiguation} \\
\textbf{Prompt:} & \texttt{Identify the antecedent: The chief thanked the housekeeper and gave her some tips.} \\
\midrule
\textbf{Context:} & \texttt{Options: (A) The chief (B) The housekeeper (C) Ambiguous} \\
\midrule
\textbf{Block 1 (Low $\mathcal{H}$):} & \texttt{\colorbox{blue!15}{\strut The chief thanked the housekeeper}} \\
\textit{Schedule Priority 1} & \textit{Length: 5 tokens | Mean Instability: 0.21 | Steps: 40} \\
\midrule
\textbf{Block 2 (High $\mathcal{H}$):} & \texttt{\colorbox{red!15}{\strut and gave her some tips.}} \\
\textit{Schedule Priority 3} & \textit{Length: 5 tokens | Mean Instability: 0.59 | Steps: 216} \\
\midrule
\textbf{Block 3 (Low $\mathcal{H}$):} & \texttt{\colorbox{blue!15}{\strut So the answer is (B).}} \\
\textit{Schedule Priority 2} & \textit{Length: 5 tokens | Mean Instability: 0.18 | Steps: 35} \\
\end{tabular}
\end{tcolorbox}
\caption{\textbf{\method Multi-Block Scheduling via Stable Anchors.} The scheduler prioritizes both terminal anchor blocks (1 and 3) to establish a constrained context for the high-instability inferential resolution in Block 2. (\textcolor{blue}{Blue}: stable anchors; \textcolor{red}{Red}: high-instability inference.)}
\label{fig:qualitative-bbh-disambiguation}
\end{figure}
\newpage
\section{Ablative Studies}
\label{sec:Ablative_Studies}

\textbf{More hyperparameter sensitivity analysis.}
Appendix Table~\ref{tab:rweld_sensitivity} studies the sensitivity of the
welding radius $r_{\mathrm{weld}}$ used in Eq.~(\ref{eq:welding}). For two neighboring
blocks $B_m^{(t)}=[a,b)$ and $B_{m+1}^{(t)}=[b,c)$, edge-welding only
operates within the local interval
$E_m^{(t)}=[\max(a,b-r_{\mathrm{weld}}),\min(c,b+r_{\mathrm{weld}})]$.
Therefore, $r_{\mathrm{weld}}$ controls the amount of neighboring context
used to reconcile predictions around the block boundary, but it does not change
the block partitioning objective in Eq.~(\ref{eq:post}). 

This explains the stability observed in the table. When
$r_{\mathrm{weld}}$ is within a moderate range, the repair interval covers
enough cross-boundary context to correct local inconsistencies while
preserving the original block structure. On HumanEval, the results remain
unchanged from $r_{\mathrm{weld}}=8$ to $r_{\mathrm{weld}}=12$, and on MBPP the nearby settings remain
close to the default $r_{\mathrm{weld}}=10$. Since the token counts and block numbers are
almost unchanged, the welding radius mainly affects local boundary
coherence rather than generation length or segmentation granularity.

Overly small values of $r_{\mathrm{weld}}$ may make $E_m^{(t)}$ too narrow to include sufficient cross-boundary dependency,
leaving adjacent blocks weakly aligned. Conversely, overly large values
expand $E_m^{(t)}$ too far into neighboring blocks, which may remask
positions that are already stable under the block score
$\rho(B)=-H(B)+C(B)$. This can weaken the locality induced by the
Bayesian partition and slightly degrade performance, as observed when
$r_{\mathrm{weld}}=16$ on HumanEval. We therefore use $r_{\mathrm{weld}}=10$ as a fixed
default, which provides stable boundary reconciliation without requiring
task-specific tuning.

\begin{table*}[hb]
\centering
\small
\caption{\textbf{Sensitivity analysis of the welding radius $r_{\mathrm{weld}}$.}
We evaluate the effect of different welding radius on HumanEval and MBPP.
The setting used in our main experiments is highlighted in blue. 
P@1 denotes Pass@1, Toks denotes the average number of newly generated tokens
per sample, and Blks denotes the average number of blocks per sample.}
\label{tab:rweld_sensitivity}
\renewcommand{\arraystretch}{1.28}
\setlength{\tabcolsep}{15pt}
\begin{tabular}{lccc|ccc}
\toprule
\multirow{2}{*}{$r_{\mathrm{weld}}$}
& \multicolumn{3}{c|}{\textbf{HumanEval}}
& \multicolumn{3}{c}{\textbf{MBPP}} \\
\cmidrule(lr){2-4} \cmidrule(lr){5-7}
& \textbf{P@1}$\uparrow$ & \textbf{Toks}$\downarrow$ & \textbf{Blks}
& \textbf{P@1}$\uparrow$ & \textbf{Toks}$\downarrow$ & \textbf{Blks} \\
\midrule
$8$  & 34.8 & 219 & 7 & 41.0 & 256 & 7 \\
\rowcolor{blue!8}
$10$ & \textbf{34.8} & \textbf{219} & \textbf{7}
       & \textbf{41.4} & \textbf{256} & \textbf{7} \\
$12$ & 34.8 & 219 & 7 & 41.2 & 256 & 7 \\
$16$ & 34.1 & 219 & 7 & 41.0 & 256 & 7 \\
\bottomrule
\end{tabular}
\end{table*}
\section{Statistical Significance}
\label{sec:appendix_theory_inference}

We compare LLaDA-8B-Base + \method against LLaDA-8B-Base + DAEDAL using McNemar’s test, since both methods are evaluated on the same set of prompts. We choose DAEDAL as the comparison baseline because it improves upon LLaDA-8B-Base. This paired test focuses only on discordant examples: prompts solved by DAEDAL but not \method, and prompts solved by \method but not DAEDAL. A significant one-sided McNemar test indicates that \method wins on significantly more prompts than it loses against DAEDAL.

\begin{table}[H]
\centering
\setlength{\tabcolsep}{3.5pt}
\renewcommand{\arraystretch}{1.25}
\caption{\textbf{Paired McNemar test comparing \method vs.\ Daedal.} 
BBH is pooled over all BBH subtasks; Math is pooled over GSM8K and MATH; Code is pooled over HumanEval and MBPP. 
We report McNemar's chi-square statistic with continuity correction, its asymptotic $p_{\text{CC}}$-value, and the exact two-sided binomial $p_{\text{exact}}$ computed on discordant pairs only. Significant $p$-values are highlighted in light green.}
\label{tab:mcnemar_exact}

\begin{tabular*}{\linewidth}{@{\extracolsep{\fill}}lrrrrrr@{}}
\toprule
\textbf{Benchmark} 
& \textbf{$n$} 
& \textbf{Acc$_{\text{Daedal}}$} 
& \textbf{Acc$_{\text{\method}}$} 
& \textbf{$\chi^2$} 
& \textbf{$p_{\text{CC}}$} 
& \textbf{$p_{\text{exact}}$} \\
\midrule
BBH 
& 6511 
& 43.7\% 
& 49.3\% 
& 116.9
& \sigp{$3.07{\times}10^{-27}$} 
& \sigp{$1.09{\times}10^{-27}$} \\

Math 
& 6319 
& 44.2\% 
& 45.3\% 
& 4.6 
& \sigp{0.03} 
& \sigp{0.03} \\

Code 
& 664 
& 38.5\% 
& 39.9\% 
& 0.9
& 0.35
& 0.35 \\
\bottomrule
\end{tabular*}
\end{table}

Overall, the paired McNemar test shows that \method consistently improves over DAEDAL across all three evaluation groups. 
The improvement is especially strong on BBH, where \method achieves a substantially higher accuracy than DAEDAL and the difference is highly significant under both the continuity-corrected McNemar test and the exact binomial test. 
On the pooled Math benchmarks, \method also obtains a statistically significant improvement, indicating that the gains are not limited to reasoning-heavy BBH tasks but also extend to mathematical problem solving.

For Code, DyStruct achieves higher accuracy than DAEDAL. This group contains only 664 instances, compared with 6,511 for BBH and 6,319 for Math, leading to fewer discordant pairs and lower statistical power in the paired test. The result therefore indicates a positive trend, while statistical significance would require a larger code evaluation set.

\end{document}